\documentclass[conference]{IEEEtran}
\usepackage{times}

\usepackage[numbers]{natbib}
\usepackage{multicol}
\usepackage[bookmarks=true]{hyperref}

\usepackage{color,soul}
\usepackage{graphicx}
\usepackage{siunitx}
\usepackage{bbold}
\usepackage{subcaption}
\usepackage{dsfont}
\usepackage{xspace}

\usepackage{multicol}
\usepackage{varwidth}

\usepackage{algorithm}
\usepackage{algpseudocode}
\usepackage{amssymb}

\usepackage[usenames,dvipsnames,table]{xcolor}

\newcommand{\algoName}{GTI\xspace}
\newcommand{\algoFull}{Generalization Through Imitation\xspace}

\title{GTI: Learning to Generalize Across Long-Horizon Tasks from Human Demonstrations}

\author{%
Ajay Mandlekar$^{*}$,
Danfei Xu$^{*\dagger}$,
Roberto Mart\'{i}n-Mart\'{i}n,
Silvio Savarese,
Li Fei-Fei\\ \\
Stanford Vision and Learning Lab \\
\thanks{ $^{*}$ Equal contribution, alphabetical order}%
\thanks{ $^{\dagger}$ Correspondence to \href{mailto:danfei@cs.stanford.edu}{danfei@cs.stanford.edu}}%

}
\IEEEoverridecommandlockouts
\begin{document}


\maketitle

\begin{abstract}
Imitation learning is an effective and safe technique to train robot policies in the real world because it does not depend on an expensive random exploration process. However, due to the lack of exploration, learning policies that generalize beyond the demonstrated behaviors is still an open challenge. We present a novel imitation learning framework to enable robots to 1) learn complex real world manipulation tasks efficiently from a small number of human demonstrations, and 2) synthesize new behaviors not contained in the collected demonstrations. 
Our key insight is that multi-task domains  often present a latent structure, where demonstrated trajectories for different tasks intersect at common regions of the state space.
We present \algoFull (\algoName), a two-stage offline imitation learning algorithm that exploits this intersecting structure to train goal-directed policies that generalize to unseen start and goal state combinations. 
In the first stage of \algoName, we train a stochastic policy that leverages trajectory intersections to have the capacity to compose behaviors from different demonstration trajectories together.
In the second stage of \algoName, we collect a small set of rollouts from the unconditioned stochastic policy of the first stage, and train a goal-directed agent to generalize to novel start and goal configurations. We validate \algoName in both simulated domains and a challenging long-horizon robotic manipulation domain in real world. Additional results and videos are available at \url{https://sites.google.com/view/gti2020/}. 
\end{abstract}

\IEEEpeerreviewmaketitle


\section{Introduction}
\label{s_intro}

Imitation Learning (IL) is a promising paradigm to train physical robots on complex manipulation skills by replicating behavior from expert demonstrations~\cite{pomerleau1989alvinn, krishnan2018learning}. However, IL suffers from an important limitation: it is difficult for the robot to generalize to new behaviors that are different from the demonstrated expert trajectories. Thus, the performance of IL depends on providing expert demonstrations that cover a wide variety of situations~\cite{ross2011reduction}. Expecting this kind of coverage from a fixed set of expert demonstrations is often unrealistic, especially for long-horizon multi-stage manipulation tasks (e.g. setting up a table or cooking), due to the combinatorial nature of possible task instances and valid solutions.

One way to generalize from a fixed amount of demonstrations is to modulate policy behaviors with a task specification, e.g., instructions~\cite{codevilla2018end}, video demonstrations~\cite{xu2018neural,huang2019neural,duan2017one}, or goal observations~\cite{chang2019learning,lynch2019learning}.
The hope is that by training the policy conditioned on the task specification, the policy can exhibit new behaviors by sampling new task specifications. 
However, due to the often complex correlation between behavior and task specification, these methods require large amounts of annotated demonstrations~\cite{lynch2019learning} and consequently, they do not scale well to physical robots in the real world.

\begin{figure}[t]
\centering
\begin{subfigure}[b]{0.25\linewidth}
\centering
\includegraphics[width=.99\linewidth]{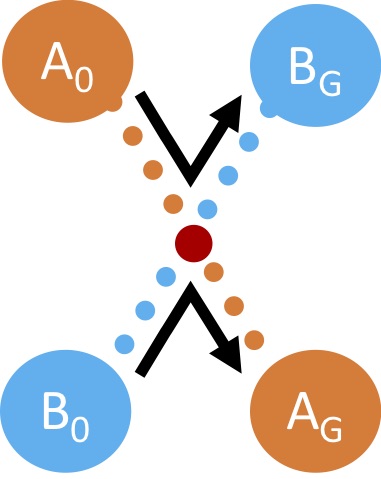}
\vspace{15pt}
\end{subfigure}
\hspace{5pt}
\begin{subfigure}[b]{0.6\linewidth}
\includegraphics[width=.99\linewidth]{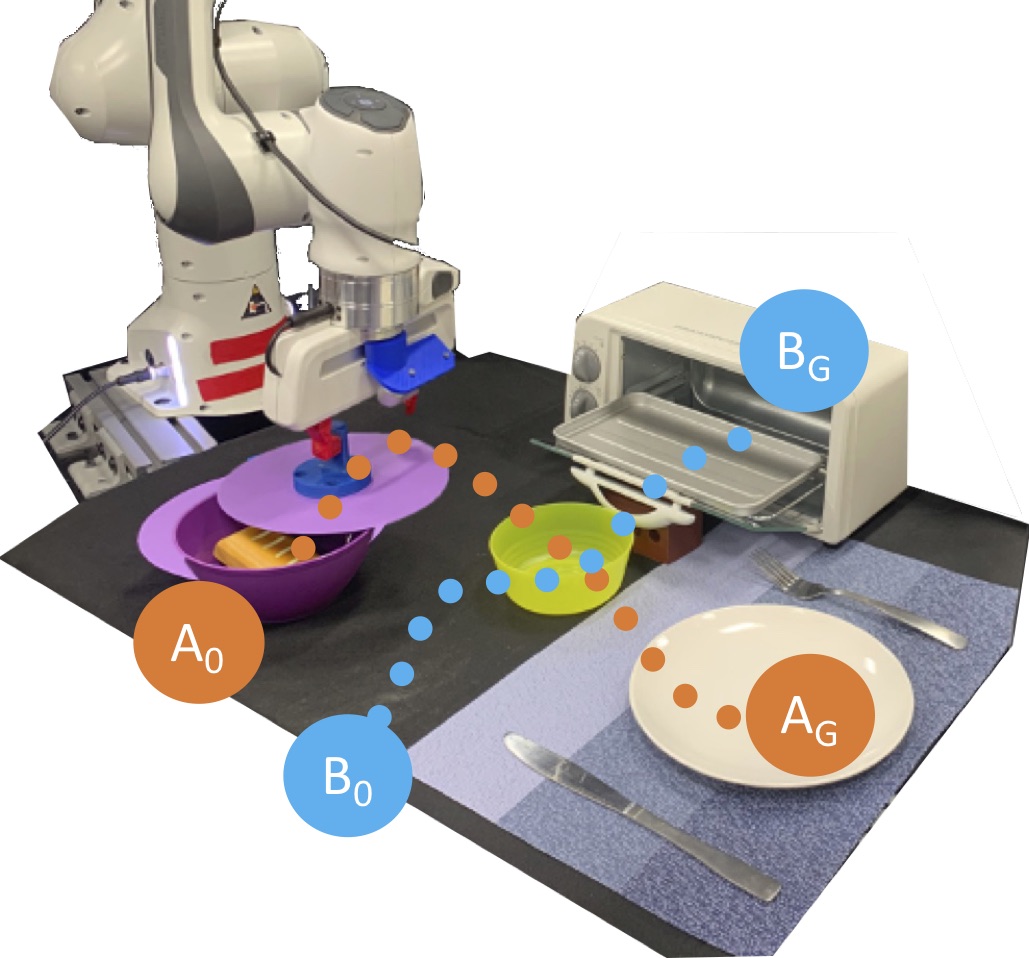} 
\end{subfigure}
\caption{\textbf{Generalizing across long-horizon tasks with intersectional structure}. (\textit{left}) Simplified diagram of an intersectional structure: The demonstrations of task A (orange) and of task B (blue) intersect at a state. Our method \textbf{Generalization Through Imitation} (GTI) generalizes to unseen behaviors (black arrows) by exploiting the compositional structure of demonstrations with intersection points. (\textit{right}) Demonstrations of long-horizon manipulation tasks in real world often present intersectional structure: demonstrations of task A (robot takes the bread from the closed container and serves it) and task B (robot picks up the bread from the table and reheats it in the oven) intersect at a state where the bread is in the bowl and the robot picked up the bowl. \algoName leverages such intersections to compose different demonstration sequences into new task trajectories.}
\label{fig:pull}
\end{figure}

\begin{figure*}[!t]
   \centering
    \includegraphics[width =0.95\linewidth]{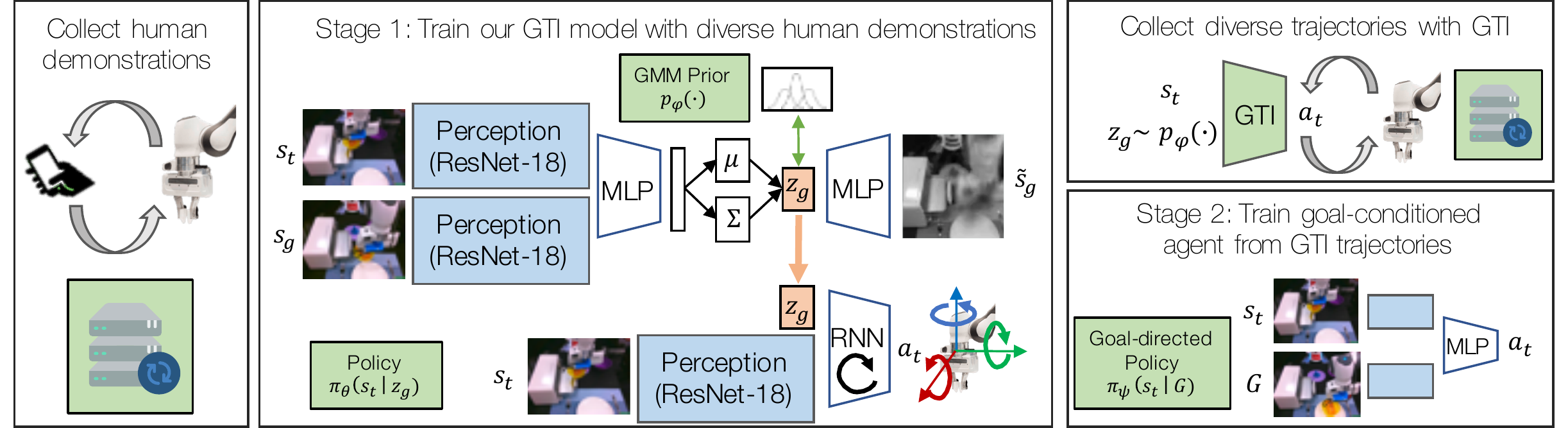}
    \caption{\textbf{\algoName Overview:} The figure summarizes our training pipeline for \algoName. We first collect human demonstrations with teleoperation using the RoboTurk interface~\cite{mandlekar2018roboturk, mandlekar2019roboturk}. Then, we train an agent on the demonstrations (\algoName Stage 1) that learns a cVAE (top of Stage 1 box) that models the distribution of future observations $s_g$ conditioned on a current observation $s_t$, and a goal-conditioned low-level policy (bottom of Stage 1 box) that learns to reach future observations from current observations with closed-loop visuomotor control. The low-level policy is conditioned on latents from the cVAE. Then, we collect rollouts from this trained agent and use them to train an goal-conditioned agent (\algoName Stage 2) that can solve novel start and goal combinations. During rollouts, the cVAE prior is used to sample latents for the low-level policy to follow. This encourages the policy to visit novel goal states from start states by leveraging trajectory intersections.}
   \label{fig:system}
\end{figure*}

We propose a method that generalizes from a few set of real demonstrations based on a key insight: in the real world, many task domains contain a structure like the one depicted in Fig.~\ref{fig:pull}, where diverse sequences of actions that start and terminate at different regions of the state space ``intersect'' at a certain state. That means that independently of the initial configuration (bread in the container or on the table) and the goal of the task (serve it on the plate or reheat it in the oven) there is a point in the sequence of actions where the state of the environment is the same (bread is in the bowl). 
In this way, it should be possible to leverage such an intersection to compose known sequences of trajectories into novel, unseen paths from one task configuration to another.

In this work, we propose \algoFull (\algoName), a method to learn novel, goal-oriented self-generated behaviors that result from the composition of different phases of a small set of demonstrations on a real physical robot. \algoName is based on visual data acquired during human teleoperation of a robot arm. Our method is comprised of two stages: a first stage of self-generating diverse and new behaviors by encouraging generalization in unintentional (random) rollouts of an imitation policy, and a second stage of learning goal-directed policies from these rollouts to achieve controllable new behaviors.
For the first stage, instead of a naive imitation of all demonstrations that collapses into a single dominant demonstration, our proposed approach derives a stochastic policy that generates diverse behaviors by making divergent choices at trajectory intersections. At its core, this variability is achieved by a generative model in the space of visual observations with a Gaussian mixture model as a prior, encouraging multimodality in the rollouts.

The contributions of our work are as follows:
\begin{itemize}
    \item We propose \algoFull (\algoName) - a novel algorithm for compositional task generalization based on learning from a fixed number of human demonstrations of long horizon tasks.

    \item We present a real world end-to-end imitation learning system that performs complex long horizon manipulation tasks with closed-loop visuomotor control and generalizes to new pairs of start and goal task configurations.
    \item We demonstrate the effectiveness of our approach in both simulated and real world experiments and show that it is possible to learn novel and unseen goal-directed behavior on long horizon manipulation domains from under an hour of human demonstrations.
\end{itemize}

\section{Related Work}
\label{s_rw}

Imitation learning has been applied to multiple domains such as playing table tennis~\cite{kober2009learning}, Go~\cite{silver2016mastering} and video games~\cite{Ross2010ARO}, and driving autonomously~\cite{bojarski2016end,pokle2019}
In robotics, the idea of robots learning from human demonstrations have been extensively explored~\cite{Schaal1999IsIL,Ijspeert2002MovementIW,Kober2010ImitationAR,englert2018learning,Finn2017OneShotVI,Billard2008RobotPB,Calinon2010LearningAR}. Imitation learning can be used to obtain a task policy for a task either by learning a mapping from observations to actions offline (e.g., behavioral cloning, BC~\cite{bain1995framework}) or by inferring an underlying reward function to solve with RL (inverse reinforcement learning, IRL~\cite{russell1998learning}).
However, both BC and IRL require multiple demonstrations of the same short-horizon manipulation task and suffers when trying to imitate long-horizon activities even when they are composed of demonstrated short-horizon skills. In this work we present a method to leverage multi-task demonstrations of long-horizon manipulation via composition of demonstrated skills. 

Researchers have mainly approached long-horizon imitation learning in two ways: a) one-shot imitation learning (OSIL), and b) hierarchical planning with imitation. 
In OSIL the goal is to generate an imitator policy that learns to perform tasks based on a single demonstration.
The task can be represented as a full video~\cite{xu2018neural}, a sequence of image keyframes~\cite{huang2019neural}, or a trajectory in state space~\cite{duan2017one}, while the training objective can be either maximizing the likelihood of action given an expert trajectory as in \cite{duan2017one,xu2018neural,huang2019neural,pathak2018zero}, matching the distribution of demonstrations (GAIL~\cite{wang2017robust}), or following a trajectory based on learned dynamics models (AVID~\cite{smith2019avid}). 
In this work, we do not assume access to task demonstrations nor instructions at test-time; we exploit variability and compositionality within the small set of given demonstrations to generate new strategies.

The second type of approach is hierarchical planning with imitation, which entails learning a high-level planner and a low-level goal-conditioned controller. The low-level controller learns to reach subgoals specified by the high-level planner, and the planner finds a path in the space of subgoals that drives the agent towards the original task goal. By reducing the frequency of the high-level planner, these methods can cope with longer horizon tasks. %
Prior works leverage imitation~\cite{lynch2019play, mandlekar2019iris} and reinforcement~\cite{pmlr-v80-co-reyes18a} within this framework.
However, these methods do not explicitly exploit the compositional structure of the demonstrations. 
In fact, if the demonstrations contain state intersections (see Fig.~\ref{fig:pull}) the policies from these methods would collapse to the most dominant mode (as shown empirically in Fig.~\ref{fig:reach_qual}). 
We address this challenge and turn it into an opportunity to generalize to new compositional behaviors.

A way to achieve the compositionality we are aiming for is to segment the demonstrations and recompose them, by leveraging methods such as TAPS~\cite{jayaraman2018time}, or a similar approach by \citet{kipf2019compositional}, which recovers subtasks from demonstrations in the spirit of hierarchical RL~\cite{sutton1999between,bacon2017option}.
In contrast, our method does not explicitly model the temporal structure of a demonstration, which can be difficult to capture especially for real world video demonstrations.

Algorithmically, we first train a goal-agnostic agent with offline imitation learning and propose a new method to enforce diversity in its behaviors. Previous approaches have looked at how to enforce diversity in an agent's behavior~\cite{eysenbach2018,florensa2017stochastic,daniel2012hierarchical}. These approaches aim at creating different behaviors that can be learned and exploited in a hierarchical RL architecture. Instead, we used the agent's new behaviors to collect a small set of new trajectories and train new goal conditioned policies using this small dataset, reducing the amount of new interactions needed.

\section{Problem Formulation}
\label{s_problem}

We consider a robot manipulation task a sequential decision making problem and model it as a discrete-time infinite-horizon Markov Decision Process (MDP), $\mathcal{M} = (\mathcal{S}, \mathcal{A}, \mathcal{T}, R, \gamma, \rho_0)$, where $\mathcal{S}$ is the state space, $\mathcal{A}$ is the action space, $\mathcal{T}(\cdot | s, a)$, is the state transition distribution, $R(s, a, s')$ is the reward function, $\gamma \in [0, 1)$ is the discount factor, and $\rho_0(\cdot)$ is the initial state distribution. At every step, an agent observes an state $s_t$ and queries a policy $\pi$ to choose an action $a_t = \pi(s_t)$. The agent performs the action and observes the next state $s_{t+1} \sim \mathcal{T}(\cdot | s_t, a_t)$ and reward $r_t = R(s_t, a_t, s_{t+1})$. 
We augment this MDP with a set of absorbing goal states $\mathcal{G} \subset \mathcal{S}$, where $g \in \mathcal{G}$ corresponds to a specific state of the world in which the task is considered to be solved.
Every pair $(s_0, \mathcal{G})$ of initial state $s_0 \sim \rho_0(\cdot)$ and goals for a task $\mathcal{G}$ corresponds to a new task instance.

We assume access to a dataset of $N$ task demonstrations $\mathcal{D} = \{\tau_i\}_{i=1}^N$ where each demonstration is a trajectory $\tau_i = (s^i_0, a^i_0, s^i_1, a^i_1, ..., s^i_{T_i})$ that begins in a start state $s^i_0 \sim \rho_0(\cdot)$ and terminates in a final (goal) state $s^i_{T_i}=g^i$. We also assume that the dataset $\mathcal{D}$ possesses a particular structure - the demonstration trajectories \textit{intersect} at certain states.

\textbf{Definition of Trajectory Intersection}: Let $\tau_1 = (s^1_0, a^1_0, s^1_1, a^1_1, ..., s^1_{T_1})$ and $\tau_2 = (s^2_0, a^2_0, s^2_1, a^2_1, ..., s^2_{T_2})$ be two trajectories from the demonstration dataset. We say that $\tau_1$ and $\tau_2$ \textit{intersect} if $s^1_i = s^2_j$ for some $i$ and $j$.

While each trajectory in the dataset demonstrates how to reach a \textit{particular} goal $g^i$ from a start state $s^i_0$, the demonstrations implicitly contain more information about unseen start-goal pairs if leveraged through novel combinations at the intersections. For example, if two trajectories $\tau_1$ and $\tau_2$ intersect at a state $s$, then there are implicit paths from $s^1_0$ to $g^2$ and $s^2_0$ to $g^1$ in the demonstration data - even though a demonstrator never explicitly produced this trajectory. 

Our goal is to generate a policy that can solve new situations, defined as pairs $(s_0, g)$ that have not been demonstrated. To do that we will exploit trajectory intersections in the demonstration data by harnessing the variability in subsequent states after the intersection with the two-stage method explained in the next section.

\begin{algorithm*}[!t]
\caption{Train Diverse-Behavior BC Policy}
\label{alg:stage1}
\vspace{5pt}
\begin{algorithmic}[1]

\Require 
\Statex $\{E_{\phi}(s_g, s), D_{\phi}(z_g, s), p_{\phi}(z_g)\}$, $\pi_{\theta}(s \,|\, z_g)$, $\pi_{\psi}(s, g)$
\Comment{Goal cVAE encoder, decoder, and prior; Stage 1 Policy}
\State Stage 1 Training
\For{$i = 1, 2, ..., n_{\text{iter}}$} 
\State $(s_t, a_t, s_{t+1}, ..., s_{t + H - 1}, a_{t + H - 1}, s_{t + H}) \sim \mathcal{D}$ \Comment{Sample $H$-length sequence from the dataset}
\State $s_g \leftarrow s_{t+H}$ \Comment{Treat last observation as goal}
\State $\mu_g, \sigma_g = E_{\phi}(s_g, s_t)$, $z_g \sim \mathcal{N}(\mu_g, \sigma_g)$ \Comment{Encode goal observation into latent goal}
\State $\phi \leftarrow \arg\min_{\phi} ||s_g - D_{\phi}(z_g, s_t) ||_2^2 + \beta_{g} KL(\mathcal{N}(\mu_g, \sigma_g) || p_{\phi}(z_g))$ \Comment{Train cVAE to reconstruct goal observations}
\State $\hat{a}_t, \hat{a}_{t+1}, ..., \hat{a}_{t + H - 1} \leftarrow \pi_{\theta}(s_{t:t+H-1} \,|\, z_g)$ \Comment{Latent goal-conditioned action sequence prediction from RNN Policy}
\State $\theta \leftarrow \arg\min_{\theta} \sum_{t' = t}^{t + H - 1} ||a_{t'} - \hat{a}_{t'}||_2^2$ \Comment{Update policy with imitation loss}
\EndFor
\end{algorithmic}
\end{algorithm*}

\begin{algorithm}
\caption{Generate and Train from Diverse Rollouts}
\label{alg:stage2}
\begin{algorithmic}[1]
\Require 
\Statex $\pi_{\theta}(s \,|\, z_g)$, $\pi_{\psi}(s, g)$
\Comment{Stage 1 Policy; Stage 2 Policy}
\State Stage 2 Data Collection
\State $\mathcal{D}_2 \leftarrow \varnothing$ \Comment{Stage 2 Dataset}
\For{$i = 1, 2, ..., n_{\text{rollouts}}$} 
\State $s \sim \rho_0(\cdot)$ \Comment{Sample a random start state}
\For{$j = 1, 2, ..., \left \lfloor{\mathcal{H} / H}\right \rfloor $} \Comment{Collect $\mathcal{H}$-length rollouts}
\State $z_g \sim p_{\phi}(\cdot)$ \Comment{Sample a goal from cVAE prior}
\For{$k=1, 2, ..., H$}
\State $a \leftarrow \pi_{\theta}(s, z_g)$
\State $s' \leftarrow \mathcal{T}(s, a)$ \Comment{Act using Stage 1 policy}
\State $\mathcal{D}_2 \leftarrow \mathcal{D}_2 \cup (s, a, s')$ \Comment{Collect transition}
\State $s \leftarrow s'$
\EndFor
\EndFor
\State $s_g \leftarrow s$, $\mathcal{D}_2 \leftarrow \mathcal{D}_2 \cup \{s_g\}$ \Comment{Annotate goal state}
\EndFor

\State Stage 2 Training
\For{$i = 1, 2, ..., n_{\text{iter}}$} 
\State $(s_t, a_t, s_g) \sim \mathcal{D}_2$ \Comment{Sample state, action, and goal}
\State $\psi \leftarrow \arg\min_{\psi} ||a_t - \pi_{\psi}(s_t, s_g)||^2$
\EndFor
\end{algorithmic}
\end{algorithm}

\section{Method}
\label{s_method}

We propose a two-stage approach to achieve compositional generalization by extracting information from intersecting demonstrations (Fig.~\ref{fig:system}). In Stage 1, a stochastic policy is trained to reproduce the diverse behaviors in the demonstrations through multimodal imitation learning. The underlying assumption is that the distribution of human demonstrations in the dataset is multimodal, where trajectory modes in the MDP arrive at intersecting states from different start states and reach different goals from intersecting states. In Stage 2, we collect sample trajectories from the multimodal imitation agent and distill them into a single goal-directed policy. The goal-directed policy is the final result of our method and allows us to control the robot to demonstrate new behaviors, solving pairs $(s_0, G)$ of initial state and goal never demonstrated by a human. In the following we will explain each stage in detail.

\subsection*{Stage 1: Multimodal Imitation Learning to Generate Novel Behaviors}

At this stage our goal is to learn a stochastic policy that is able to 1) reproduce diverse behaviors from the demonstration dataset and 2) compose such behaviors together to produce novel, unseen trajectories in the environment. In order to properly leverage trajectory intersections, the policy needs to be able to perform \textit{compositional imitation}, where the policy imitates one trajectory before an intersection and a second trajectory after an intersection. 

We propose to decompose the imitation learning problem into two subproblems: 1) learning a generative model from the database of demonstrations to predict possible future states conditioned on a current state, and 2) training a goal-conditioned imitation policy from the demonstrations using predicted states as goals. 
The generative model resulting from 1) acts as high-level goal proposal for the low-level goal-conditioned policy that results of 2). 
This decomposition increases the interpretability of the imitation policy because it allows inspection of future states, and has been shown to improve performance over a flat imitation policy~\cite{mandlekar2019iris}. 

Using this decomposition in our task setup, we reduce the problem of generating diverse behaviors by leveraging intersecting demonstrations to the problem of training a diverse goal proposal model. Since the low-level policy is goal-conditioned, it is not affected by the multimodality in the possible goals and can be trained as a simple goal-conditioned visuomotor policy such as in IRIS~\cite{mandlekar2019iris}. A crucial difference to prior work is imposed by the need to deploy policies in the real world, where we do not have access to the ground truth state of the environment, but only to raw images. The two level policy learning hierarchy we propose for Stage 1 is designed to be compatible with high-dimensional image observations. 

The proposed training approach is described in Algorithm~\ref{alg:stage1}. Formally, we propose to learn (1) a low-level goal-conditioned controller, $\pi_{\theta}(s, s_g)$, that outputs actions to try and reach a goal observation $s_g$ that is $H$ timesteps away, and (2) a high-level goal proposal network, $p_{\phi}(s_g | s)$, that samples goals for the low-level controller to reach. $H$ is a fixed policy horizon. Both models are trained on $H$-length subsequences from demonstration trajectories. In the following we explain how each model is trained.

\textbf{Goal Proposal Network on Image Observations:} The goal proposal model has to generate possible future states conditioned on a current state. 
Our proposal model is a conditional Variational Autoencoder (cVAE)~\cite{kingma2013auto}. The cVAE learns a conditional distribution $p(s_{t+H} | s_t)$ to produce future observations that are $H$ steps away from the current observation. Since our states are represented by images, we train a cVAE on the image sequences of the demonstrations.

\begin{figure*}[!t]
\centering
\begin{subfigure}[b]{0.25\linewidth}
\centering
\includegraphics[width=.99\linewidth]{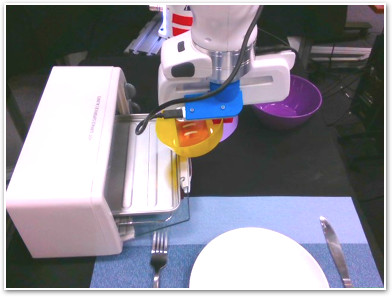}
\caption*{Current Image Observation}
\end{subfigure}
\hfill
\begin{subfigure}[b]{0.74\linewidth}
\centering
\includegraphics[width=.99\linewidth]{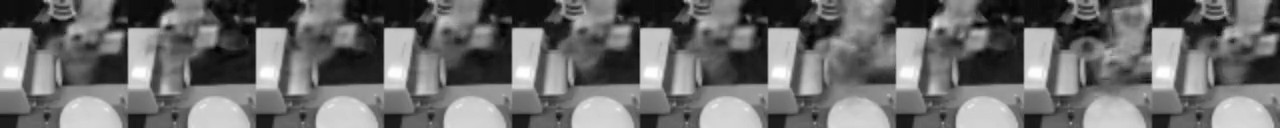}
\\
\vspace{1pt}
\includegraphics[width=.99\linewidth]{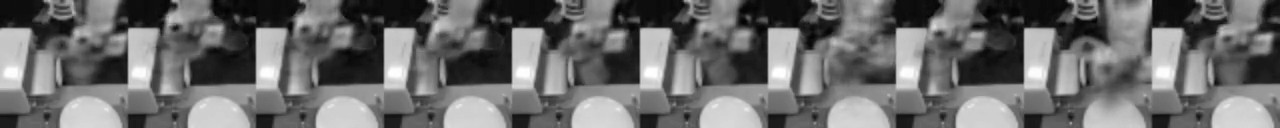}
\\
\vspace{1pt}
\includegraphics[width=.99\linewidth]{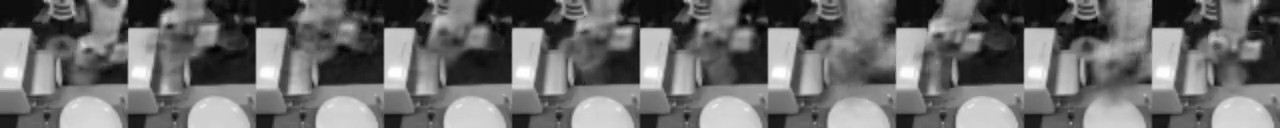}
\caption*{\begin{tabular}{cccccccccc}
~~~~1~~~~&~~~~2~~~~&~~~~3~~~~&~~~~4~~~~&~~~~5~~~~&~~~~6~~~~&~~~~7~~~~&~~~~8~~~~&~~~~9~~~~&~~~10~~~~
\end{tabular}
}
\end{subfigure}
\caption{\textbf{Qualitative cVAE Goal Observation Samples:} Our cVAE over future observations $p_{\phi}(s_{t+H} | s_t)$ is trained with a learned GMM prior. Here, we sampled 3 latents per GMM mode at a trajectory intersection point and visualize the generated images. Notice that modes 7 and 9 generate observations that correspond to serving the plate of bread while the other modes generate observations that correspond to putting the plate of bread in the oven. Also notice that there is significant variation from images generated by a single mode, corresponding to the temporal variation present in the dataset.}
\label{fig:vae_qual}
\vspace{-5pt}
\end{figure*}

Given the multimodal nature of the demonstrations in our dataset, we propose to learn a Gaussian Mixture Model (GMM) prior $p_{\phi}(z) = \sum_{k=1}^K w^k_{\phi} \mathcal{N}(\mu^k_{\phi}, (\sigma^k_{\phi})^2)$ in lieu of a standard Gaussian $\mathcal{N}(0, 1)$ prior. This flexible GMM prior resulted in diverse, multimodal image distributions, as shown in Fig.~\ref{fig:vae_qual}. 
The GMM prior plays a crucial role in overcoming some of the challenges of imitation learning for physical manipulation in the real world. For example, while we hope that robot actions are applied at a regular frequency and that observations are received periodically and without lag, the reality is that both are subject to stochastic delays due to sensor communication and motor actuation. Consequently, there can be significant variation in observations that are $H$ action steps away from the current observation. This issue is exacerbated by noisy human demonstrators that are not consistent in task execution speed. This makes multimodal goal generation even more important, so that the temporal variability within each mode of future observations can also be captured by the model. Indeed, Fig.~\ref{fig:vae_qual} shows that the trained cVAE can capture such variations in future observations.

Structurally, the encoder, decoder, and prior networks share the same architecture and weights for image feature extractors. These feature extractors terminate with a spatial-softmax layer~\cite{finn2016deep} that infer low dimensional keypoint representations from spatial feature representations. We found that this was an efficient way to downsize image features while also encouraging learning a succinct image feature representation.
We model the cVAE decoder as a Multi-Layer Perceptron (MLP) and train it to reconstruct lower resolution grayscaled images, which has been shown to produce better results for visuomotor tasks~\cite{finn2016deep}. 
The full architecture of the cVAE is shown in Fig.~\ref{fig:system} (Stage 1, top). 

\textbf{Goal-Conditioned Imitation using Latent Goals:} %
Given the goals from the proposal network, the imitation policy should generate actions to reach that goal by conditioning on the goal.
Conditioning the low-level controller on high-dimensional image observations as goals is undesirable - the model could learn spurious associations for what is important in a goal image, and VAE image reconstructions are often low quality and noisy. Instead, we condition the low-level controller on a lower dimensional latent goal that encodes the most salient features of goal observations. In order to do this, we developed a \textit{goal relabeling} scheme that leverages the latent space of the high-level cVAE to generate goals for the low-level controller. 

At train time, for a given subsequence $(s_t, a_t, s_{t+1}, a_{t+1}, ..., s_{t+H})$ sampled from the dataset of demonstrations, instead of choosing the goal $s_g = s_{t+H}$, we first use the cVAE encoder to retrieve a sample from the posterior distribution $z_g \sim \mathcal{N}(\mu_{\phi}(s_t, s_{t+H}), \sigma^2_{\phi}(s_t, s_{t+H}))$ and set $s_g = z_g$. Thus, the low-level controller is trained to produce actions $a_i = \pi_{\theta}(s_i, z_g)$ for $i \in [t, t+H-1]$. At test time, instead of using the high-level cVAE decoder to generate goals, we sample latent goals directly from the learned cVAE prior $z_g \sim p_{\phi}(\cdot)$. The goal-conditioned policy is shown in Fig.~\ref{fig:system} (Stage 1, bottom).

\begin{figure}[!t]
\begin{subfigure}[b]{0.66\linewidth}
\centering
\includegraphics[width =0.99\linewidth]{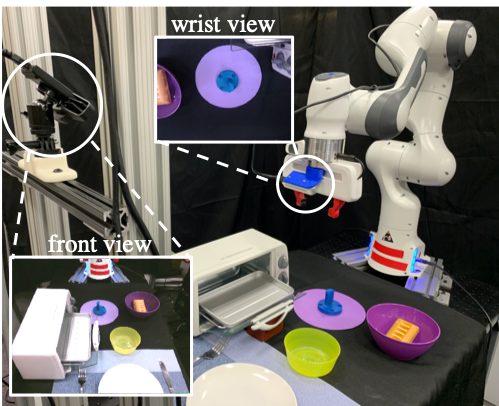}
\end{subfigure}
\hfill
\begin{subfigure}[b]{0.32\linewidth}
\centering
\includegraphics[width =0.99\linewidth]{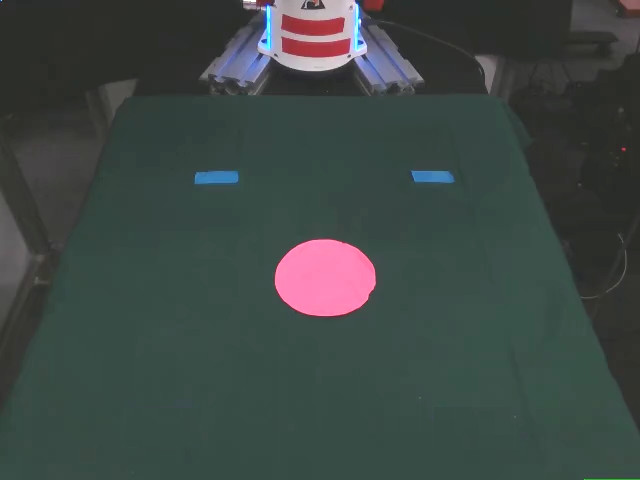}
\\
\vspace{12pt}
\includegraphics[width =0.99\linewidth]{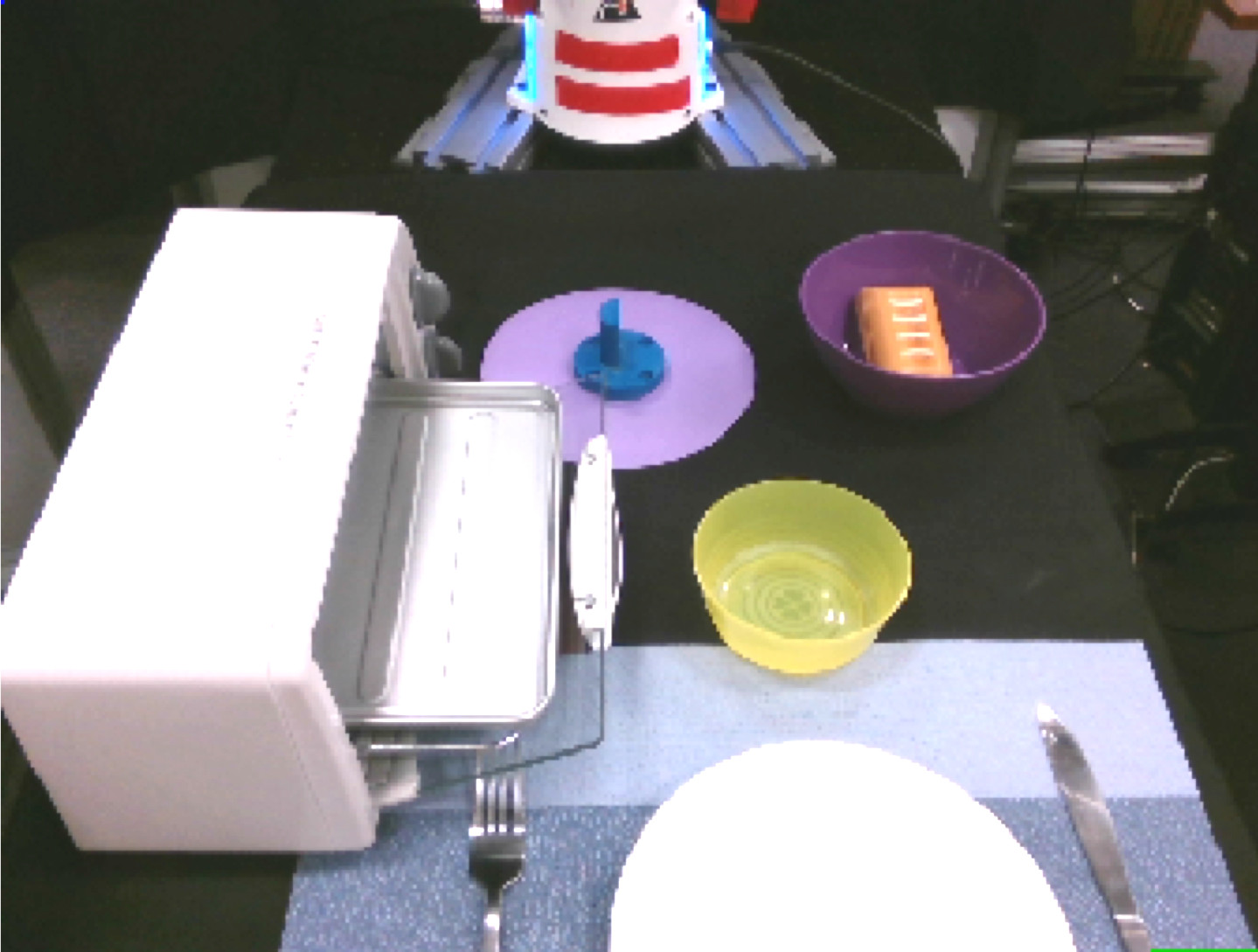}
\end{subfigure}
\caption{\textbf{Real World Robot Manipulation Setup}: (\textit{left}:) Our Panda robot mounted rigidly to a table. The manipulation is recorded with two cameras, a front-viewing camera rigidly mounted on the table and second camera mounted on robot's wrist (views from each camera depicted).
(\textit{right}:) Humans teleoperate to perform long-horizon tasks such as \texttt{PandaReach} (\textit{top}) and \texttt{PandaKitchen} (\textit{bottom}). Our goal is to generate novel behaviors by composing visual demonstrations with intersecting states.}
\label{fig:robot_setup}
\end{figure}

\subsection*{Stage 2: Goal-Directed Imitation of Novel Behaviors}

In the second stage we make use of the models trained in Stage 1 to generate new demonstrations through compositional generalization. While the goal of the first stage is to leverage trajectory intersections to obtain new, unseen behaviors, in this second stage we aim to learn to control these newly demonstrated behaviors in a goal directed manner.

The proposed approach to collect new data and train a final goal-directed policy is described in Algorithm~\ref{alg:stage2}. In this stage, we collect a set of additional trajectories with the learned stochastic policy from Stage 1, which may include new, unseen state sequences through trajectory intersections (Fig.~\ref{fig:system}, top right). Then, a goal-directed agent is trained from these trajectories in order to learn novel, goal-directed behavior (Fig.~\ref{fig:system}, bottom right). 

To train the goal-directed agent from these additional policy rollouts, we resort to goal-directed behavioral cloning. For a given rollout taken by the stage 1 stochastic policy $\tau = (s_0, a_0, s_1, a_1, ..., s_T)$, we treat the final state reached as a goal observation $g = s_T$, and train a network $\pi_{\psi}$ to predict every action conditioned on the current and goal observation, $a_t = \pi_{\psi}(s_t, g)$. Since the goal observation is held constant over an entire trajectory (as opposed to the Stage 1 agent), we did not find a need to use lower dimensional latent representations for the goal observation, and instead train directly from raw image input. At test-time, we condition the trained agent on a new goal observation, and it executes a rollout to try and reach that goal. In this way, the new behaviors exhibited by the stochastic Stage 1 policy are distilled into the Stage 2 agent that can now handle novel start and goal state combinations. 

We train our Stage 2 model using a single step goal-directed behavioral cloning method instead of hierarchical models (as in Stage 1) in order to ensure that we do not conflate the representational capacity of the Stage 2 policy with the quality of the dataset generated by the Stage 1 policy.

\begin{figure}[!t]
\centering
\begin{subfigure}[b]{0.185\linewidth}
\centering
\includegraphics[width=.99\linewidth]{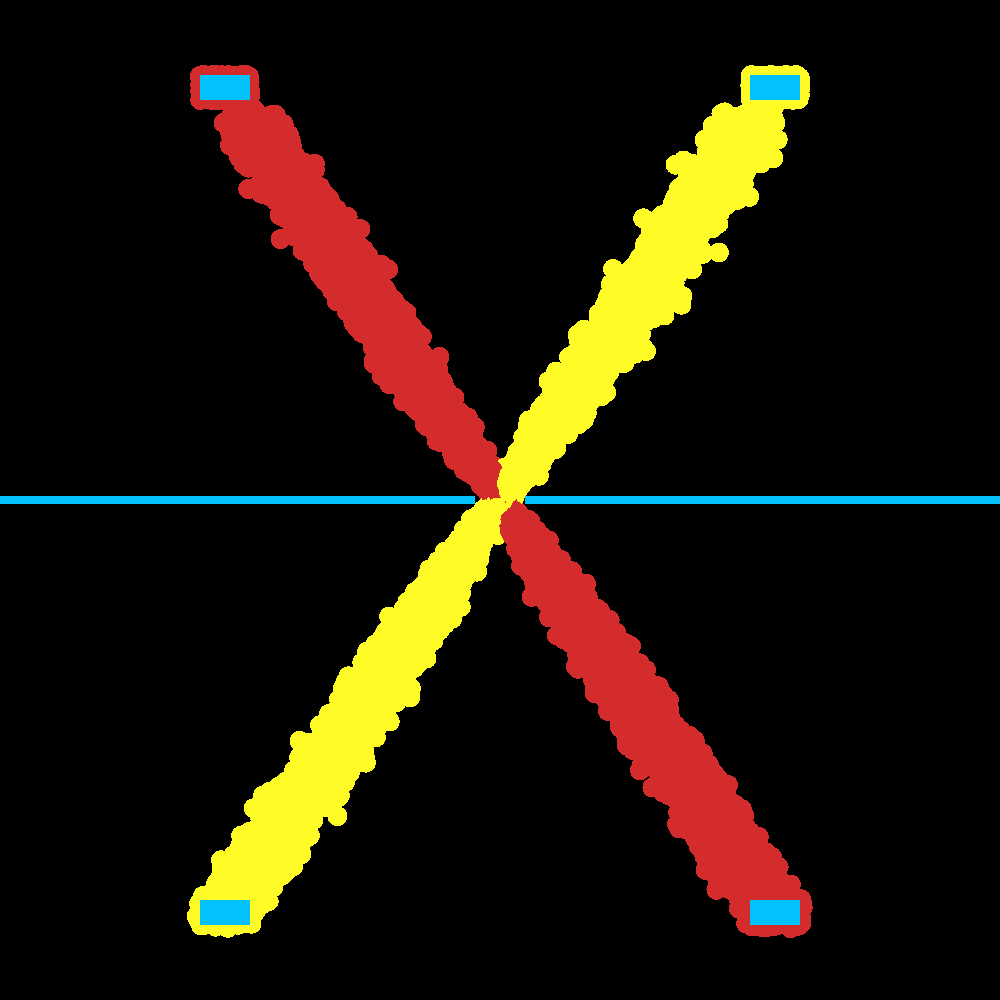}
\\
\vspace{3pt}
\includegraphics[width=.99\linewidth]{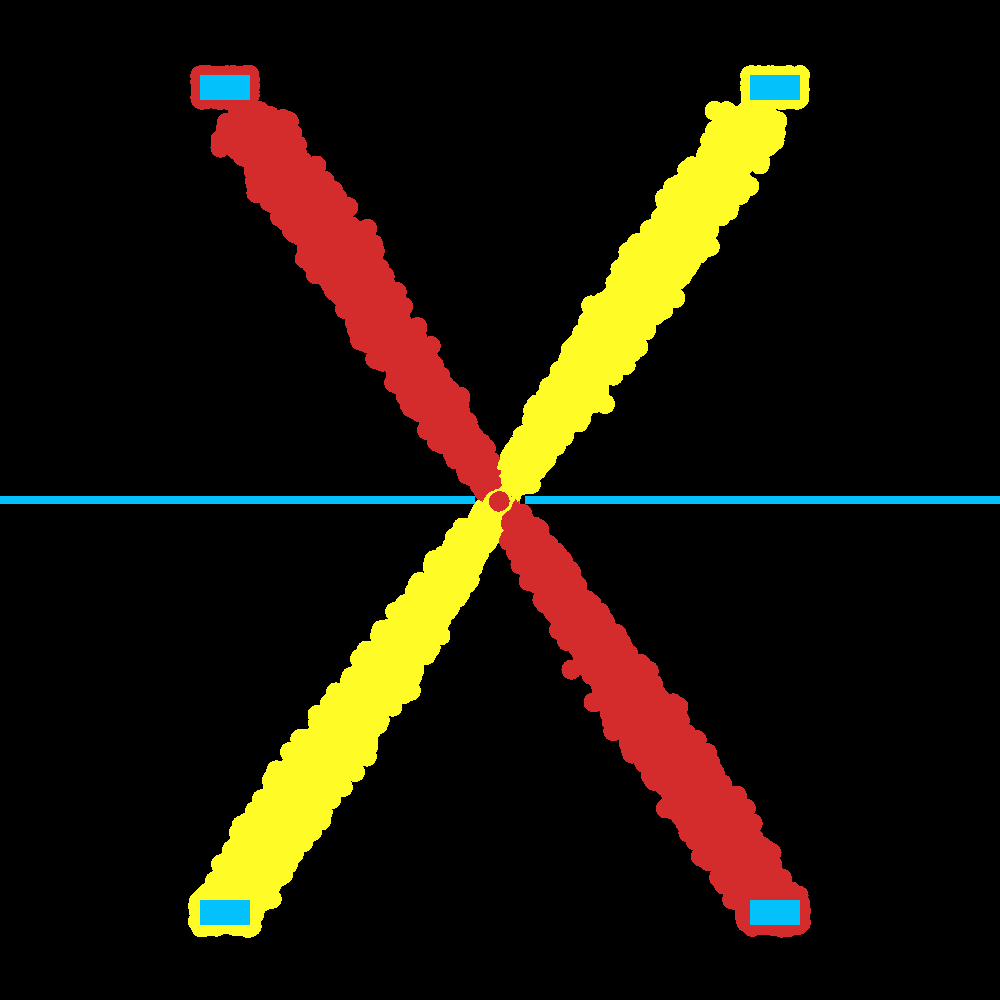}
\caption*{Dataset}
\end{subfigure}
\begin{subfigure}[b]{0.185\linewidth}
\centering
\includegraphics[width=.99\linewidth]{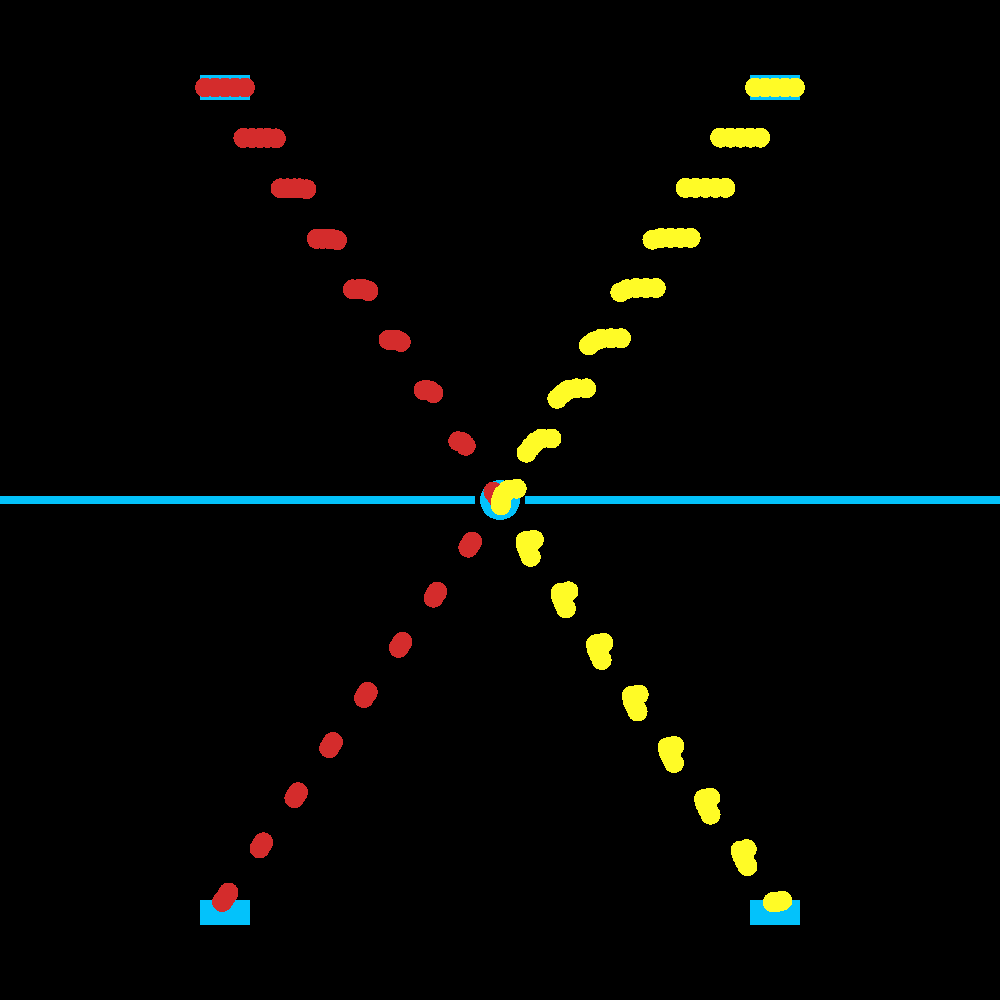}
\\
\vspace{3pt}
\includegraphics[width=.99\linewidth]{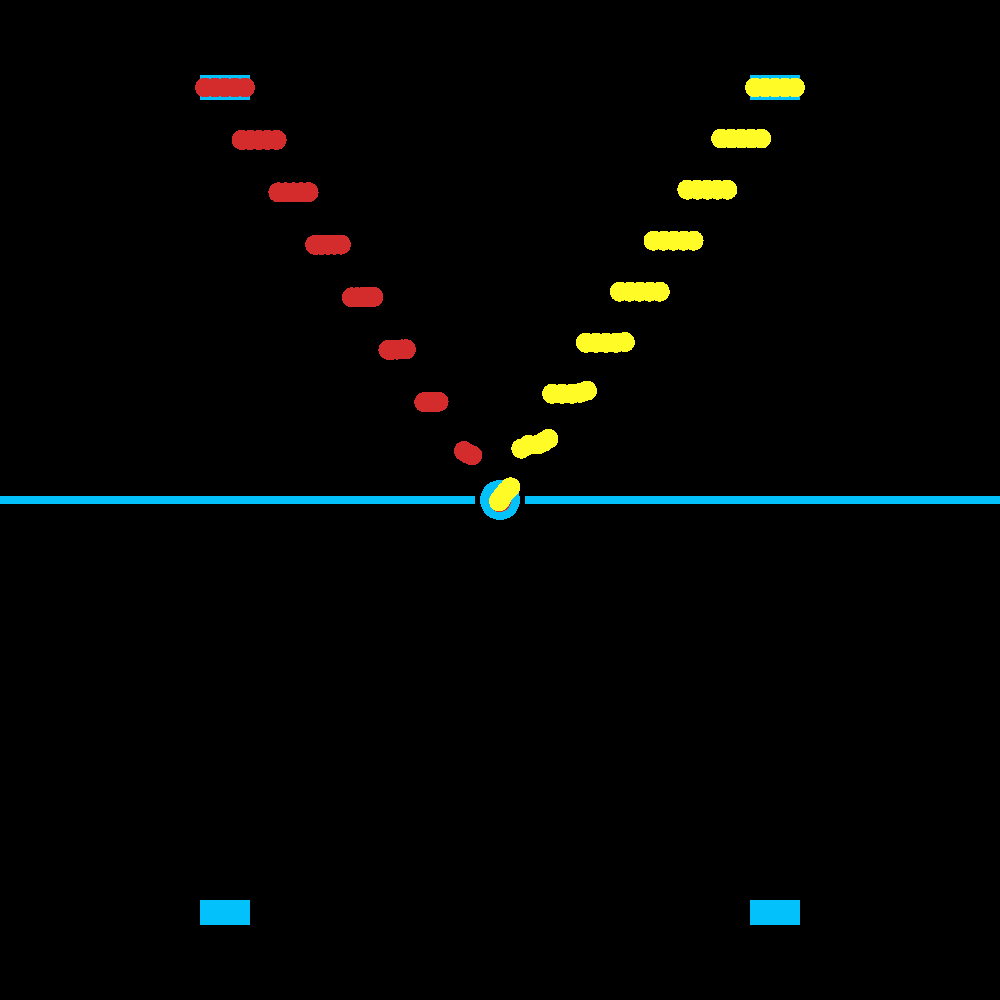}
\caption*{BC}
\end{subfigure}
\begin{subfigure}[b]{0.185\linewidth}
\centering
\includegraphics[width=.99\linewidth]{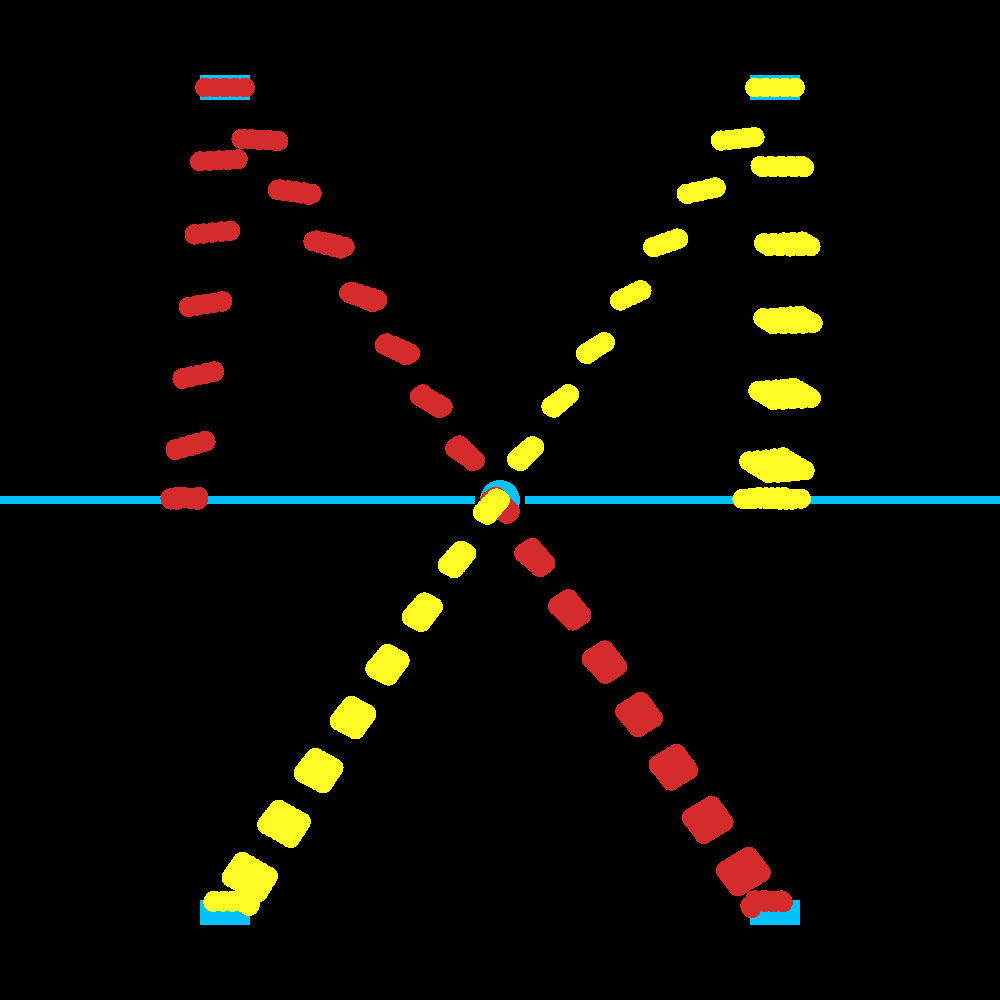}
\\
\vspace{3pt}
\includegraphics[width=.99\linewidth]{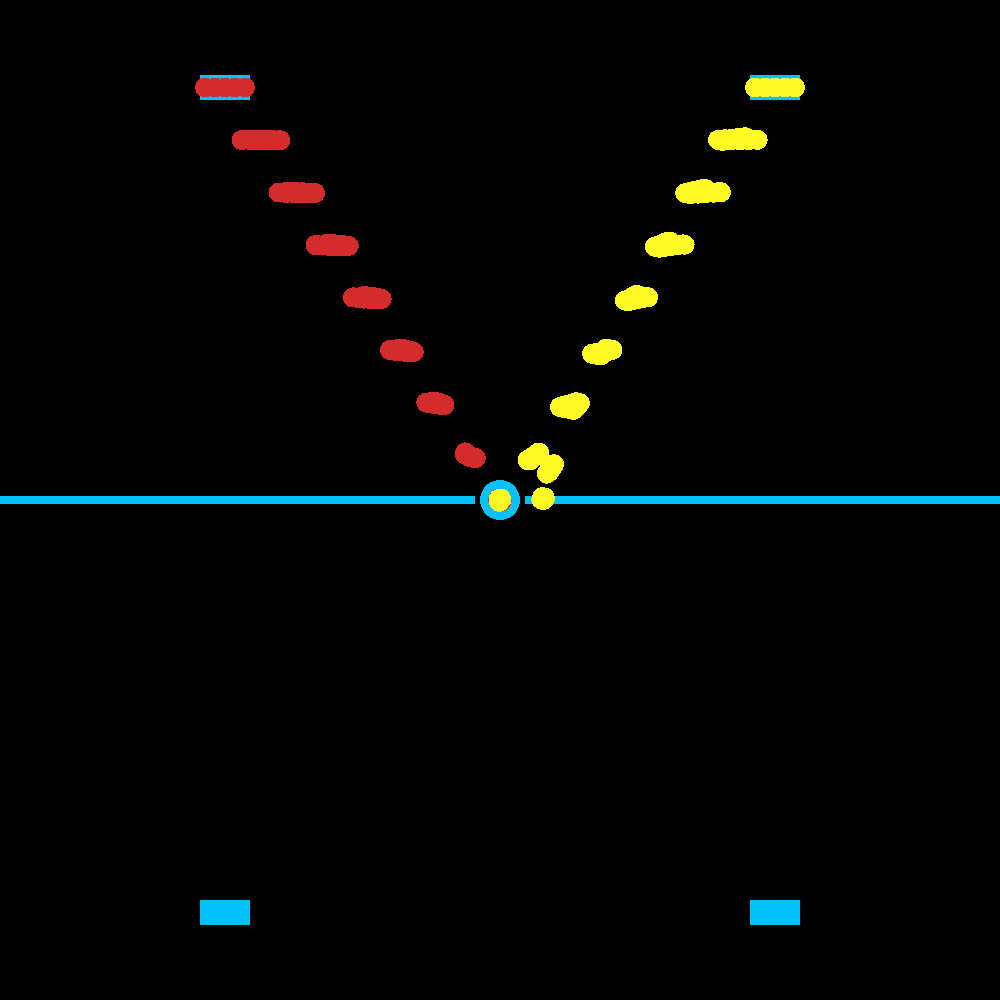}
\caption*{GCBC}
\end{subfigure}
\begin{subfigure}[b]{0.185\linewidth}
\centering
\includegraphics[width=.99\linewidth]{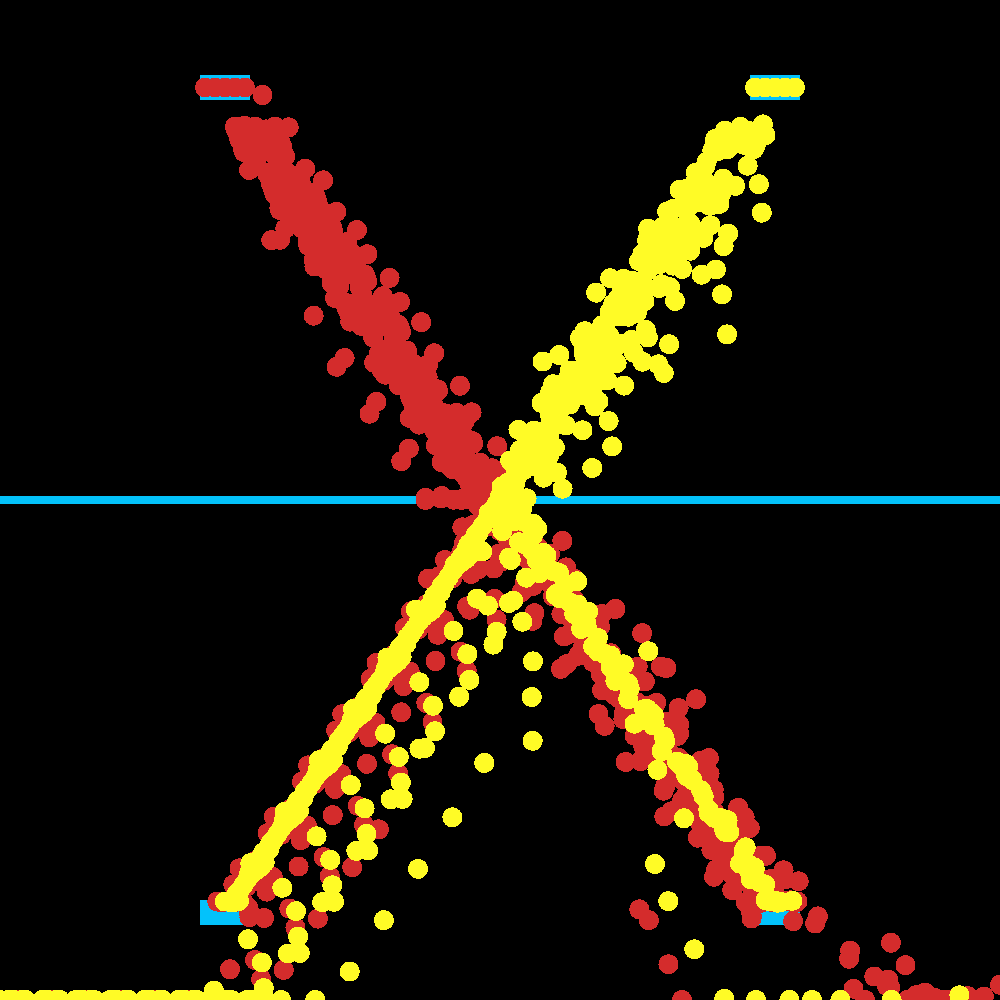}
\\
\vspace{3pt}
\includegraphics[width=.99\linewidth]{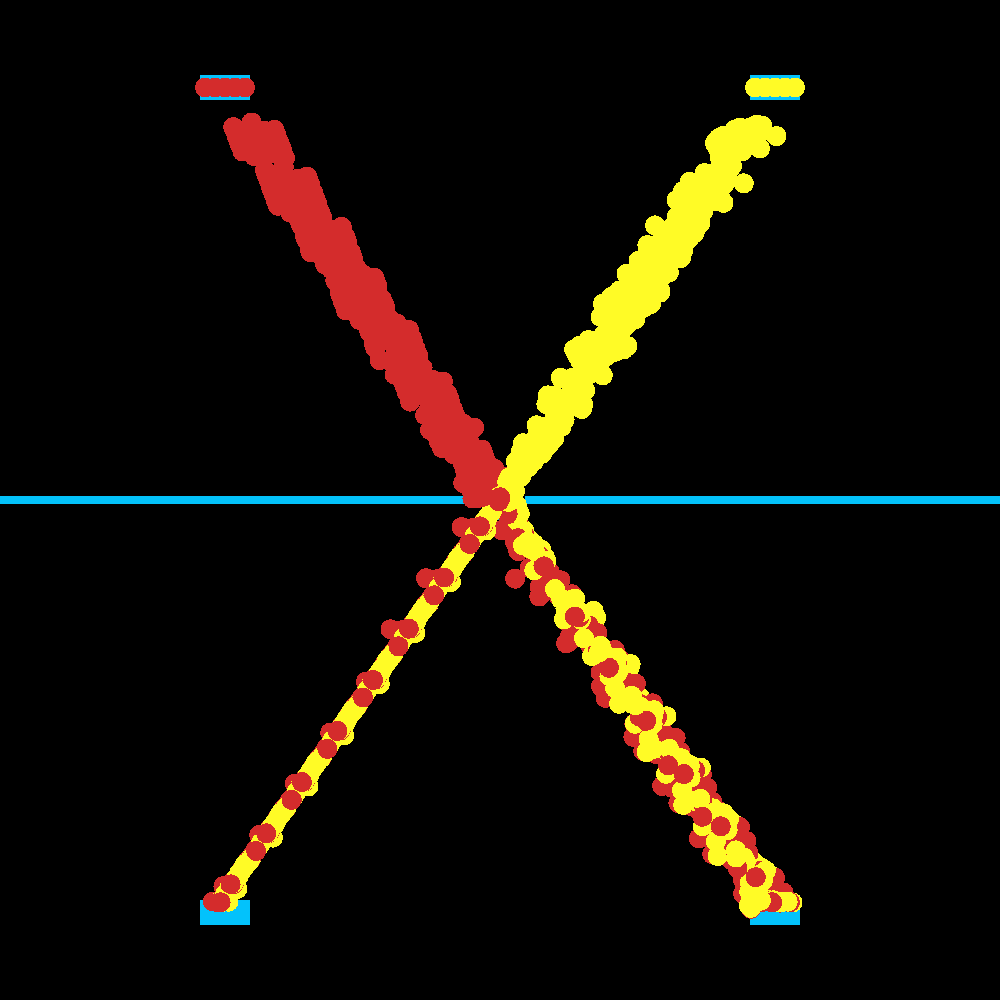}
\caption*{\algoName}
\end{subfigure}
\begin{subfigure}[b]{0.185\linewidth}
\centering
\includegraphics[width=.99\linewidth]{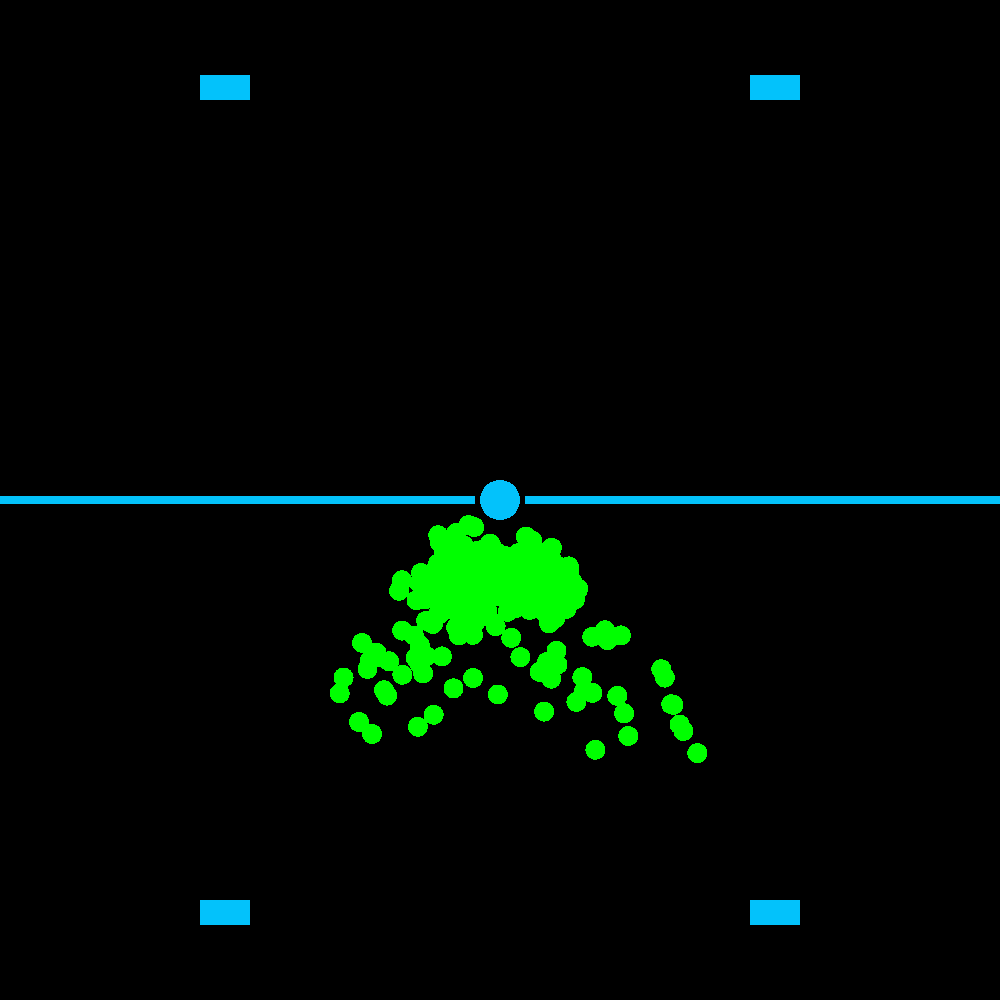}
\\
\vspace{3pt}
\includegraphics[width=.99\linewidth]{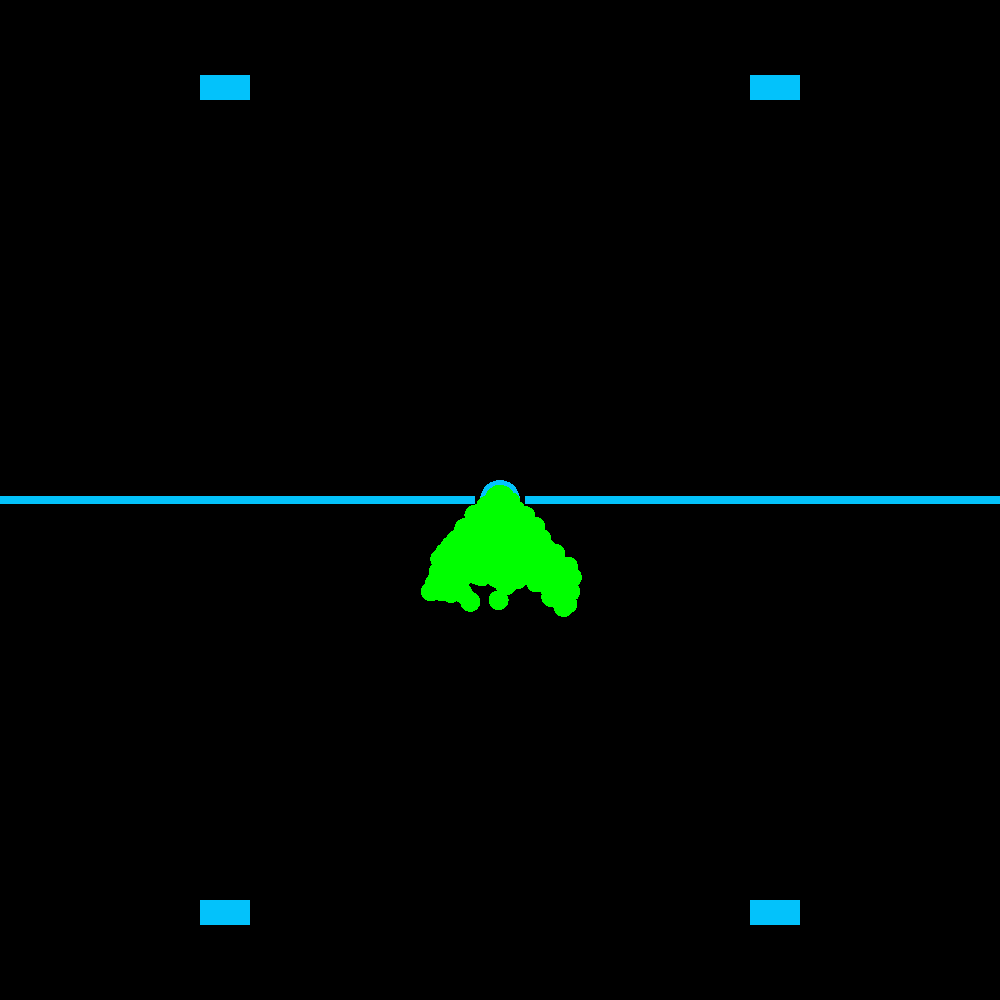}
\caption*{\algoName Goals}
\end{subfigure}
\caption{\textbf{Qualitative Stage 1 Evaluation in Simulated Domains:} The top row shows the \texttt{PointCross} domain while the bottom shows the \texttt{PointCrossStay} domain. From left to right, we show the trajectories in each dataset, policy rollouts from a trained BC model, a trained GCBC model, a trained GTI Stage 1 model (ours), and goal predictions from our cVAE at the center location. Red paths start from the top left and yellow paths start from the top right. The plots demonstrate that our method is the only one able to both reproduce original behavior and generate new behavior unseen in the demonstration data by modeling different potential outputs in the goal space.}
   \label{fig:sim_qual}
\end{figure}

\begin{figure*}[t!]
\includegraphics[width=\linewidth]{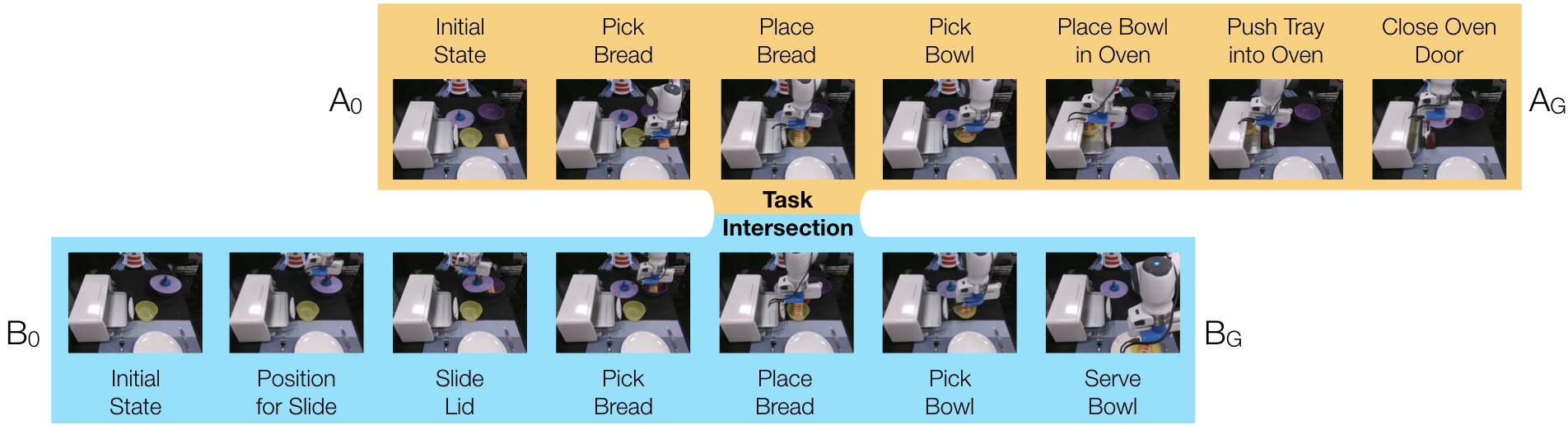}
\caption{\textbf{\texttt{PandaKitchen} Domain}: This domain consists of two initial task configurations (bread-on-table and bread-in-bowl) and two goal configurations (bread-in-oven and bread-on-plate). Each task consists of several stages, as shown in the diagram. Demonstrations are collected for the orange and blue task sequences. The diagram also shows the task intersection - in this way it is possible for a policy to leverage different demonstration sequences to solve novel start and goal combinations. Qualitative result videos are available at \url{https://sites.google.com/view/gti2020/}.}
\centering
\label{fig:pandakitchen}

\end{figure*}
\section{Experiments}
\label{s_experiments}

We evaluated \algoName in two sets of experiments. In the first set we evaluate \algoName in a simulated environment in order to have access to ground truth states and perform an in-depth analysis without confounding sources of error (e.g. inaccuracies in execution of robot commands, sensor delays, visual observations, etc). In the second set of experiments, we test \algoName in our target setup: generalization from a small number of demonstrations in real-world long-horizon tasks.

We compare \algoName against two other baselines: Behavioral Cloning (BC) and Goal-Conditioned Behavioral Cloning (GCBC). Behavioral Cloning takes state-action pairs $(s_t, a_t)$ from the demonstrations and regresses actions on states, $a_t = \pi_{\theta}(s_t)$. We expect that BC will collapse to a single mode in the demonstrations at trajectory intersections, since there is no variability captured in action outputs. GCBC is identical to the model trained in \algoName in Stage 2: we train a BC model where the input is augmented with the final observation in each demonstration, $a_t = \pi_{\theta}(s_t, s_T)$. We expect that GCBC will work well for start and goal pairs that have been covered in the demonstration data, but will fail to generalize to new pairs of start states and goals. 

\subsection{Simulation Experiments}

The purpose of our simulation tasks is to understand both quantitatively and qualitatively how well Stage 1 policies can leverage trajectory intersections to both imitate trajectories from the dataset and produce new trajectories. In our simulated domains, we train using low-dimensional observations, and consequently do not use latent goals. We evaluate two simulated tasks described below.

\textbf{\texttt{PointCross}}:
We developed a pedagogical task in a 2D simulated navigation domain where an agent begins each episode at a start location and must navigate to a goal location. The state space is 2D - the $x$ and $y$ location of the agent on the grid, and the action is also 2D - the delta $x$ and delta $y$ movement of the agent. The agent starts each episode in one of two regions - the upper left or upper right blue squares shown in the top left square of Fig.~\ref{fig:sim_qual}. The agent must navigate to either the lower left or lower right blue squares. The environment also has a narrow bottleneck in the middle that the agent must pass through to reach the bottom half of the domain from the top half. We collected 1000 noisy demonstrations in this domain using a hardcoded policy. All demonstration trajectories go from the top left to the bottom right or the top right to the bottom left. Thus, this domain mirrors our pedagogical example of trajectory intersection in Fig.~\ref{fig:pull} - there is no demonstration showing how to go from a start location in the top left square to the bottom left square, or from the top right square to the bottom right square.

\textbf{\texttt{PointCrossStay}}:
This task is in the same domain as \texttt{PointCross}, but during each episode, the policy we used for collection stayed in the middle of the grid for a random period of time before continuing on towards the goal. This biased the distribution of demonstration actions towards staying near the middle of the grid instead of continuing on to a goal, making it important for agents that learn from the dataset to model full distributions of outcomes at the middle of the grid.

\subsubsection*{Evaluation of Simulation Experiments}
In Table~\ref{table:sim} we report results on both the \texttt{PointCross} task and \texttt{PointCrossStay} task across BC, GCBC, and \algoName. We evaluate each policy over a uniform grid of 10 start locations - 5 in the upper left, and 5 in the upper right. For each start location, we collect 100 rollouts. For GCBC, we condition 50 of the rollouts on a goal in the lower left region and 50 on a goal in the lower right region.

We measure 4 different metrics averaged over all start locations and rollouts: (1) the Goal Reach Rate, which corresponds to the percentage of rollouts that actually reach the lower left or lower right corners, (2) the Seen Behavior metric, which corresponds to the percentage of goal-reaching rollouts that start and end on opposite sides of the y-axis (indicating similar behavior to demonstrations), (3) the Unseen Behavior metric, which is the percentage of goal-reaching rollouts that start and end on the same side of the y-axis (indicating novel behavior), and (4) the Occupancy metric, which is 100\% for a start location if it contains rollouts that reach goals on both the lower left and lower right, 50\% if it only reaches a single one, and 0\% if no goals are reached.

We also present visualizations of the simulation tasks, policy rollouts, and \algoName goal visualizations at the bottleneck in Fig.~\ref{fig:sim_qual}. On the \texttt{PointCross}, BC is able to consistently reach goal locations 100\% of the time, but it collapses to exactly one goal location per start location and is unable to reach both goal locations from a single start location, resulting in 50\% occupancy. The top second from the left panel of Fig.~\ref{fig:sim_qual} shows that BC collapses to a single mode depending on what direction the agent enters. On the other hand, GCBC excels at reproducing start and goal combinations from the training data, but completely fails on start and goal combinations that are unseen (top left to bottom left and top right to bottom right), resulting in 50\% success rate, no unseen behavior, and the paths depicted in the top middle column of Fig.~\ref{fig:sim_qual}. By contrast, our method is able to generate both unseen and seen behavior with a high success rate. It is able to produce trajectories that reach both the lower left and lower right goals from both the top left and top right starting states, as shown in the second from the right top panel of Fig.~\ref{fig:sim_qual}.

On the \texttt{PointCrossStay} task, neither BC nor GCBC is able to reach any goal states, because of the conflicting action supervision at the origin, since most actions keep the agent there. By contrast, our method is able to model diverse future states near the origin due to the cVAE and GMM prior, and understand how to move there with the low-level controller. Thus, by randomly sampling future states with the cVAE and having the low-level controller try to reach them, the \algoName agent is able to escape the origin and reach both goal locations from both start locations.

\subsection{Real World Robot Manipulation Experiments}

Fig.~\ref{fig:robot_setup} depicts our physical robot workspace. Our workspace is a rig that consists of a Franka Emika Panda robotic arm, a front-view Intel RealSense SR300 camera, and a wrist-mounted Intel RealSense D415 camera. Both the robot arm and the front-view camera are rigidly attached to the table. The wrist-mounted camera points towards the grasping area in front of the fingers. During teleoperation, we collected RGB images from both the front-view camera and the wrist-mounted camera at approximately \SI{20}{\hertz}. 

We collected task demonstrations by teleoperating the robot arm with full 6-DoF end effector control of the arm. We leverage the RoboTurk robot teleoperation interface~\cite{mandlekar2018roboturk, mandlekar2019roboturk} to collect task demonstrations from humans. To control the robot, a human demonstrator moves their smartphone in free space to control the robot. The 6D pose of the phone serves as target for an operational space controller~\cite{khatib1987unified} that outputs the robot joint torques to minimize the distance to the target in Cartesian space. The action space for the controller policy is 7-dimensional: delta end effector position (3-dimensional), delta orientation (3-dimensional in Euler angle representation), and a binary open/close gripper command (1-dimensional). The predicted action is transformed into end effector target for controlling the robot with the operational space controller.

The low-level policy in Stage 1 and the goal-conditional policy in Stage 2 share the same model architecture: A ResNet-18 network followed by a spatial-softmax layer~\cite{finn2016deep}. The policies take in both front-view and wrist-view RGB images as input. The images are concatenated channel-wise before being fed into the network. The high-level goal proposal network takes only the front-view images as input. The cVAE latent dimension is set to be 2.

In our experiments we perform two real world robot manipulation tasks with different levels of complexity, that we describe in the following.

\textbf{\texttt{PandaReach}}
The robot arm starts at a central location, and the goal is to reach to a table location that is either on the left or the right of the robot from the front viewing perspective (see Fig.~\ref{fig:robot_setup}, right top). We collected 20 trajectories for each goal location and hope to recover a policy that is able to visit each goal location equally without conditioning on a particular goal location. In other words, the policy needs to exhibit multimodal decision making capability. This task is a simple case to evaluate the capabilities of the cVAE in \algoName to generate multimodal predictions and policy rollouts.

\begin{figure}[t]
\centering
\begin{subfigure}[b]{0.49\linewidth}
\centering
\includegraphics[width=.99\linewidth]{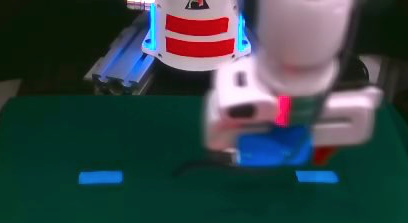}
\\
\begin{tabular}{l r}
Left=0 & Right=40\\ \hline
\end{tabular}
\caption{BC}
\end{subfigure}
\hfill
\begin{subfigure}[b]{0.49\linewidth}
\centering
\includegraphics[width=.99\linewidth]{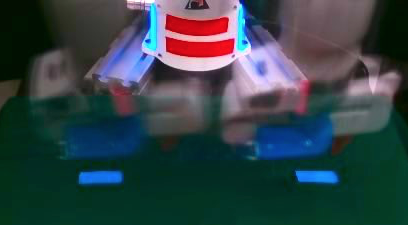}
\\
\begin{tabular}{c c}
Left=17 & Right=23\\ \hline
\end{tabular}
\caption{Ours}
\end{subfigure}
\caption{\textbf{Panda Reach Results} Toy reaching task to evaluate the ability of each model to reach multiple goals. We omit GCBC since providing perfect goal information at test-time is enough for a model to reach either the left or right target location. We present the average image of the final observation of 40 rollouts for BC and our method. BC collapses to visiting a single goal location consistently while our method is able to visit both goal locations with nearly equal visitation.}
\label{fig:reach_qual}
\end{figure}

\begin{table}[t]
\centering
\caption{\textbf{Quantitative Stage 1 Evaluation on Simulated Domains:} Performance of BC, GCBC, and \algoName, in demonstrating a combination of seen and unseen behavior in order to reach both goal locations from both start locations.}
\begin{tabular}{|c|cccc|}
\hline
{\color[HTML]{000000} Task} & \multicolumn{4}{c|}{PointCross} \\ \hline
Model & \begin{tabular}[c]{@{}c@{}}Goal Reach \\ Rate\end{tabular} & \begin{tabular}[c]{@{}c@{}}Seen \\ Behavior\end{tabular} & \begin{tabular}[c]{@{}c@{}}Unseen\\ Behavior\end{tabular} & Occupancy\\ \hline
BC & 100\% & 0.0\% & 100\% & 50.0\%    \\
GCBC & 50.0\% & 100\% & 0.0\% & 50.0\%   \\
GTI (ours) & 77.2\% & 68.9\% & 31.1\% & 100\% \\ \hline
\hline
\hline
{\color[HTML]{000000} Task} & \multicolumn{4}{c|}{PointCrossStay} \\ \hline
Model & \begin{tabular}[c]{@{}c@{}}Goal Reach \\ Rate\end{tabular} & \begin{tabular}[c]{@{}c@{}}Seen \\ Behavior\end{tabular} & \begin{tabular}[c]{@{}c@{}}Unseen\\ Behavior\end{tabular} & Occupancy\\ \hline
BC & 0.0\% & 0.0\% & 0.0\% & 0.0\%    \\
GCBC & 0.0\% & 0.0\% & 0.0\% & 0.0\%   \\
GTI (ours) & 97.2\% & 52.0\% & 48.0\% & 100\% \\ \hline
\end{tabular}
\label{table:sim}
\vspace{-8pt}
\end{table}

\begin{table}[t]
\caption{\textbf{Quantitative Evaluation on Panda Kitchen Task:} Success rate (SR) of BC, GCBC, and \algoName in reproducing demonstrated  and novel behavior in undirected setting (Stage 1) and goal-directed setting (Stage 2)}
\centering

\begin{tabular}{|c|cccc|}
\hline
{\color[HTML]{000000} } & \multicolumn{4}{c|}{Stage 1} \\ \hline
Model & \begin{tabular}[c]{@{}c@{}}A\textsubscript{0}\\ SR\end{tabular} & \begin{tabular}[c]{@{}c@{}}A\textsubscript{0}\\ Generalization\end{tabular} & \begin{tabular}[c]{@{}c@{}}B\textsubscript{0}\\ SR\end{tabular} & \begin{tabular}[c]{@{}c@{}}B\textsubscript{0}\\ Generalization\end{tabular} \\ \hline
BC & 55.0\% & 0.00\% & 40.0\% & 10.0\% \\
GCBC & - & - & - & - \\
GTI (ours) & \textbf{71.4\%} & \textbf{50.0\%} & \textbf{46.7\%} & \textbf{42.0\%} \\ \hline
\hline
\hline
{\color[HTML]{000000} } & \multicolumn{4}{c|}{Stage 2} \\ \hline
Model & \begin{tabular}[c]{@{}c@{}}A\textsubscript{0} to A\textsubscript{G}\\ SR\end{tabular} & \begin{tabular}[c]{@{}c@{}}B\textsubscript{0} to B\textsubscript{G}\\ SR\end{tabular} & \begin{tabular}[c]{@{}c@{}}\textbf{A\textsubscript{0} to B\textsubscript{G}}\\ \textbf{SR}\end{tabular} & \begin{tabular}[c]{@{}c@{}}\textbf{B\textsubscript{0} to A\textsubscript{G}}\\ 
\textbf{SR}\end{tabular} \\ \hline
BC & - & - & - & - \\
GCBC & \textbf{60.0\%} & 50.0\% & 0.00\% & 0.00\% \\
GTI (ours) & 50.0\% & \textbf{70.0\%} & \textbf{80.0\%} & \textbf{50.0\%} \\ \hline
\end{tabular}
\label{table:real}
\vspace{-15pt}
\end{table}
\textbf{\texttt{PandaKitchen}}
This domain consists of a set of complex long-horizon tasks in a cooking setup (see Fig.~\ref{fig:pandakitchen}). We collected two kinds of trajectories, each corresponding to a start and goal state pair. In the first kind of demonstration, a loaf of bread starts on the table. We denote this start configuration as $A_0$. The robot grasps the bread and places it into a yellow bowl. Next it takes the bowl and places it into the oven to reheat the bread. This requires the robot to first place the bowl into the oven, then push the sliding tray closed, and then finally close the oven door. This corresponds to the robot reheating the bread in the oven. We denote this goal configuration as $A_G$. In the second kind of demonstration, the loaf of bread starts in a large purple bowl that is covered by a lid. We denote this start configuration as $B_0$. The robot must slide the lid off of the purple bowl, grasp the loaf of bread, place the bread into the yellow bowl, and then grab the bowl and place it onto the white plate. This corresponds to the robot retrieving the loaf of bread from a covered container to serve it to someone. We denote this goal configuration as $B_G$. Thus, demonstrations are provided on the $A_0$ to $A_G$ and $B_0$ to $B_G$ tasks.

We would like the robot to generalize to two unseen start and goal configurations: (1) the robot should be able to pick the loaf off the table (instead of from the covered container) and still serve it on the plate ($A_0$ to $B_G$), and (2) the robot should be able to retrieve the bread from the covered container and place it into the oven for cooking ($B_0$ to $A_G$).

\subsubsection*{Evaluation of Real Robot Experiments}

We first discuss results on our \texttt{PandaReach} setup. Fig.~\ref{fig:reach_qual} demonstrates that our Stage 1 policy is able to visit both goal states with nearly equal visitation, while BC collapses to visiting exactly one goal state consistently. This agrees with our simulation results and suggests that our Stage 1 policy learning method can also successfully exhibit multimodal behavior on high-dimensional image observations.

Next, we present results on our \texttt{PandaKitchen} setup. We first evaluate our trained \algoName Stage 1 policy against BC to understand how well each method can perform undirected imitation to reach both kinds of goal configurations from both kinds of start configurations. The left column of Table ~\ref{table:real} reports the success rate of each policy from each start configuration - which is the percentage of rollouts that end in either $A_G$ or $B_G$, and the generalization percentage of each start configuration - which is the percentage of successful rollouts that result in a unseen goal configuration ($A_0$ to $B_G$ or $B_0$ to $A_G$). 

We found that the Stage 1 \algoName policy is able to solve 71.4\% of $A_0$ task instances, where the bread begins on the table, and visits both $A_G$ and $B_G$ goal configurations equally, ensuring diverse rollouts. The success rate on $B_0$ task instances, where the bread begins inside the covered container, is $46.7\%$, which is lower, but successful rollouts end in novel goal configurations $42\%$ of the time. We hypothesize that the lower success rate is due to the long-horizon nature of the $B_0$ to $A_G$ task, where the robot must first uncover the container and retrieve the bread, then drop it in the bowl, place the bowl inside the oven, push the oven tray inside, and finally close the oven door. 

To train our goal-directed Stage 2 policy for \algoName, we collected rollouts from the Stage 1 undirected policy until 50 successful demonstrations had been collected for each start configuration. We evaluate our trained \algoName Stage 2 goal-directed policy against GCBC on all four start and goal combinations to understand how well our method can reproduce train-time trajectories and generalize to new start and goal combinations. This comparison showcases the value of the diverse \algoName Stage 1 rollouts: our Stage 2 policy is a GCBC model trained on Stage 1 rollouts, while the baseline is the same GCBC model trained on the original set of demonstrations. The right column of Table~\ref{table:real} reports the success rate of all 4 start and goal combinations. For this goal-directed evaluation, we collect 20 rollouts per start and goal pair, for each model. Every rollout begins by conditioning the model on a random goal configuration. For example, for the $B_0$ to $A_G$ evaluation, the workspace is reset to a configuration where the loaf of bread is inside the container, and the model is provided with an image observation where the robot has closed the oven door successfully, with the loaf of bread inside.

We found that the goal-directed \algoName policy results in significantly higher success rates than the GCBC model trained on the original set of demonstrations. It is able to achieve a $50\%$ success rate or higher on all 4 start and goal combinations - including the novel $A_0$ to $B_G$ and $B_0$ to $A_G$ tasks. This result is significant when considering that the Stage 2 \algoName policy was solely trained on the set of rollouts generated by the Stage 1 \algoName policy, suggesting that goal-directed imitation of self-generated diverse rollouts provides a stronger supervision signal than learning directly from the source demonstrations. This validates our approach of utilizing the source demonstrations to model short, diverse behaviors from the dataset, using a learned, undirected policy to produce novel trajectories using the learned diverse behaviors, and finally learning goal-oriented behavior from these policy rollouts.

\section{Conclusion}
\label{s_conclusion}

Common robotic manipulation tasks and domains possess intersectional structures, where trajectories through intersect at different states. In this work, we presented \algoFull (\algoName), a novel algorithm to achieve compositional task generalization from a set of task demonstrations by leveraging such trajectory crossings to generalize to unseen combinations of task initializations and desired goals. We demonstrated that \algoName is able to both reproduce behaviors from demonstrations and, more importantly, generalize to novel start and goal configurations, both in simulated domains and a challenging real world kitchen domain.
There are many avenues for future work. Leveraging \algoName in the context of unstructured ``play'' data~\cite{lynch2019play} or large scale crowdsourced robotic manipulation datasets~\cite{mandlekar2019roboturk} is a promising direction, as there are likely to be many trajectories that intersect at several locations in the state space. Demonstrations of humans exploring an environment contain multiple opportunities to leverage intersections for compositional task generalization in order to learn new goal-directed skills with \algoName.


{
\section*{Acknowledgment}
Ajay Mandlekar acknowledges the support of the Department of Defense (DoD) through the NDSEG program. We acknowledge the support of Toyota Research Institute (``TRI''); this article solely reflects the opinions and conclusions of its authors and not TRI or any other Toyota entity. We acknowledge ONR grant N00014-19-1-2477 and ONR MURI grant N00014-16-1-2127.
}

\bibliographystyle{plainnat}
\bibliography{references}


\end{document}